\documentclass[10pt,twocolumn,letterpaper]{article}

\usepackage{cvpr}
\usepackage{times}
\usepackage{epsfig}
\usepackage{graphicx}
\usepackage{amsmath}
\usepackage{amssymb}

\usepackage{multirow}
\usepackage{color}
\usepackage{amsmath}
\usepackage{algorithm,algpseudocode}

\newcommand{\ourapproach}{Depth-wise Decomposition}

\newcommand{\x}[1]{$#1\times$}

\newcommand{\ratio}{speed-up ratio}

\newcommand{\sepconv}{depth-wise separable convolution}

\newcommand{\Tensordecom}{Tensor factorization}

\newcommand{\channelpruning}{channel pruning}

\newcommand{\structured}{structured simplification}

\newcommand{\implementation}{optimized implementation}
\newcommand{\Implementation}{Optimized implementation}
\newcommand{\sparseconnect}{connection pruning}

\DeclareMathOperator*{\argmin}{arg\,min}
\newcommand{\aeq}[1]{  \begin{equation}
      \begin{aligned}
    #1
      \end{aligned}
  \end{equation}}

\newcommand{\approach}{depth-wise decomposition}

\usepackage[pagebackref=true,breaklinks=true,letterpaper=true,colorlinks,bookmarks=false]{hyperref}

\cvprfinalcopy 

\def\cvprPaperID{3} 

\begin{document}

\title{Depth-wise Decomposition for Accelerating Separable Convolutions\\ in Efficient Convolutional Neural Networks}

\author{
Yihui He$^{1}$, Jianing Qian$^2$, Jianren Wang$^3$, Cindy X. Le$^4$, Congrui Hetang$^5$, Qi Lyu$^6$, Wenping Wang$^7$, Tianwei Yue$^8$ \\
$^1$Carnegie Mellon University; \texttt{he2@alumni.cmu.edu} \\
$^2$Carnegie Mellon University; \texttt{jianingq@alumni.cmu.edu} \\
$^3$Carnegie Mellon University; \texttt{jianrenw@alumni.cmu.edu} \\
$^4$Columbia University; \texttt{xl2738@columbia.edu} \\
$^5$Carnegie Mellon University; \texttt{congruihetang@gmail.com} \\
$^6$Michigan State University; \texttt{lyuqi1@msu.edu} \\
$^7$Carnegie Mellon University; \texttt{wenpingw@alumni.cmu.edu} \\
$^8$Carnegie Mellon University; \texttt{tyue@alumni.cmu.edu} \\
}
\maketitle

\begin{abstract}
Very deep convolutional neural networks (CNNs) have been firmly established as the primary methods for many computer vision tasks. However, most state-of-the-art CNNs are large, which results in high inference latency. Recently, depth-wise separable convolution has been proposed for image recognition tasks on computationally limited platforms such as robotics and self-driving cars. Though it is much faster than its counterpart, regular convolution, accuracy is sacrificed. In this paper, we propose a novel decomposition approach based on SVD, namely \approach, for expanding regular convolutions into depth-wise separable convolutions while maintaining high accuracy. We show our approach can be further generalized to the multi-channel and multi-layer cases, based on Generalized Singular
Value Decomposition (GSVD)~\cite{gsvd}.
We conduct thorough experiments with the latest ShuffleNet V2 model~\cite{shufflenetv2} on both random synthesized dataset and a large-scale image recognition dataset: ImageNet~\cite{imagenet}. Our approach outperforms channel decomposition~\cite{cd} on all datasets. More importantly, our approach improves the Top-1 accuracy of ShuffleNet V2 by \textbf{$\sim$2\%}. 
\end{abstract}

\section{Introduction}
In recent years, very deep convolutional neural networks (CNNs)~\cite{zf,xception,mnasnet} have led to a series of breakthroughs in many image understanding problems~\cite{imagenet,coco,ADE20K,vg}, such as image recognition~\cite{resnet,googlenet}, object detection~\cite{fast,faster,softer,klloss,zhu2019feature}, semantic segmentation~\cite{fcn,hypercolumn,lightweight, wang2019prediction}. Most state-of-the-art CNNs have very high inference latency, although outperforming the previous approaches.
However, on computational budgets limited platforms such as smartphones, wearable systems~\cite{wang2018vertical,xia2017validation}, surveillance cameras~\cite{camera}, and self-driving cars~\cite{djuric2018motion}, visual recognition tasks are required to be carried out in a timely fashion. 

\begin{figure*}
    \begin{center}
      \includegraphics[width=0.9\linewidth]{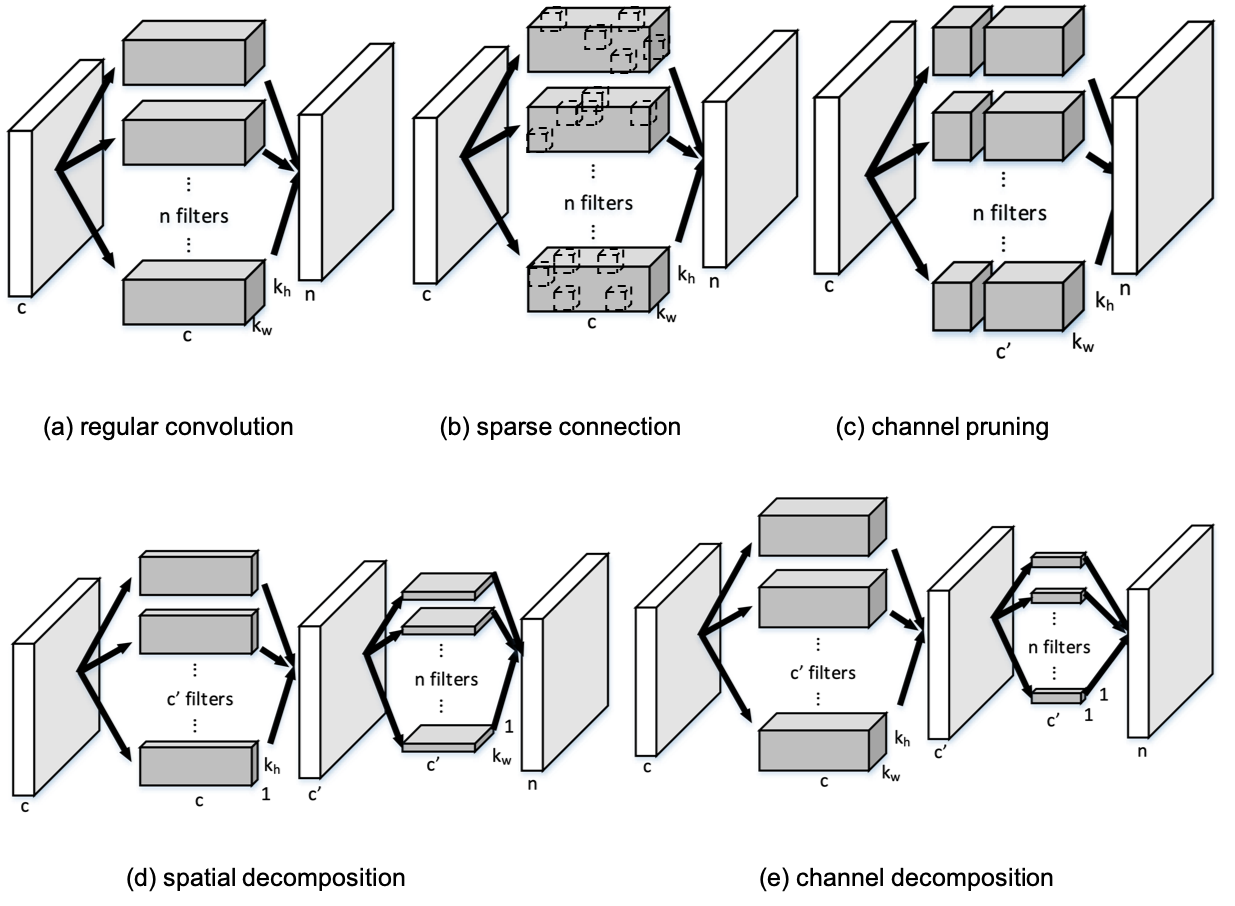}
    \end{center}
    \caption{Different types of compression algorithms}
    \label{fig:related}
\end{figure*}

Driven by the increasing need for faster CNN models, research focus has been moving towards reducing
CNN model size and computational budgets while achieving acceptable accuracy instead of purely pursuing very high accuracy.  For example, MobileNets~\cite{mobilenets} proposed a family of lightweight convolutional neural networks based on depth-wise separable convolution. ShuffleNets~\cite{shufflenet,shufflenetv2} proposed channel shuffle to reduce parameters and FLOPs (floating point operations per second). To further decrease parameters, ShiftNet~\cite{shiftnet} proposed shifting feature maps as an alternative to spatial convolution. The other trend is to compressing large CNN models, shown in Figure~\ref{fig:related} (a). For example, channel decomposition~\cite{cd} and spatial decomposition~\cite{jaderberg2014speeding} proposed to decompose regular convolutional layers, shown in Figure~\ref{fig:related} (e) and (d). Channel pruning~\cite{cp} proposed to prune channels of convolutional layers, shown in Figure~\ref{fig:related} (c). Deep compression~\cite{han2015deep,lit} proposed to sparsify the connections of fully-connected layers, shown in Figure~\ref{fig:related} (b).

All the recent approaches that try to design compact CNN models mentioned above share a common component: \sepconv, initially introduced by \cite{sepconv}. As is pointed out in MobileNets~\cite{mobilenets}, $3\times3$ depth-wise separable convolutions use between \textbf{8 to 9} times less computation than standard convolutions, with the same number of input and output channels. However, inevitably the accuracy is also sacrificed. Although many efforts have been put on searching optimal hyper-parameters for the compact models~\cite{nasnet,mnasnet}, it remains a question \textbf{whether we could improve the performance of \sepconv\ itself}.

Motivated by this, in this paper, we take \textbf{the first step} towards mitigating the performance degradation of \sepconv. Inspired by channel decomposition work~\cite{cd} which decomposes a standard CNN to a narrower CNN without much degradation (shown in Figure~\ref{fig:related} (e)), we propose to decompose an "unseparated" convolution into a \sepconv, while minimizing the performance degradation. Specifically, we first replace all \sepconv s in a compact convolutional neural network (\eg, ShuffleNet V2~\cite{shufflenetv2}) with standard convolutions. Then the standard convolutional neural network is trained from scratch on ImageNet. Finally, we decompose the standard convolutions into \sepconv\ to obtain a new compact CNN which is of the same architecture however performs better. 

To show the generality of our approach, we conduct rich experiments on ShuffleNet V2 and Xception with the large-scale ImageNet dataset. Using random data, our approach is comparable to channel decomposition~\cite{cd}. For single layer \textbf{\x{9}} acceleration, our approach consistently performs better than channel decomposition~\cite{cd} for different convolutional layers. For decomposition of the ShuffleNet V2~\cite{shufflenetv2} whole model, our approach achieves \textbf{$\sim$ 2\%} better Top-1 accuracy than the original ShuffleNet V2.

We summarize our contributions as follow:
\begin{enumerate}
    \item We propose a novel decomposition approach, namely \ourapproach, for expanding regular convolution into \sepconv.
    \item We take the first step towards improving \sepconv\ performance itself, to the best of our knowledge.
    \item Our proposed method improves the Top-1 accuracy of the original model by around 2\%.
\end{enumerate}

\section{Related Work}
Since LeCun \etal\ introduced optimal brain damage~\cite{lecun1989optimal,hassibi1993second}, there has been a significant amount of works on accelerating CNNs~\cite{cheng2017survey}. Many of them fall into several categories: designing efficient architectures~\cite{mobilenets,shufflenetv2}, \implementation~\cite{bagherinezhad2016lcnn}, quantization~\cite{Rastegari2016}, and \structured~\cite{jaderberg2014speeding}.

\subsection{Designing Efficient Architecture}
Depth-wise separable convolution is widely used in efficient networks like MobileNets~\cite{mobilenets}, ShuffleNets~\cite{shufflenet} and MnasNets~\cite{mnasnet}. \cite{nd} proved that a regular convolution could be approximated by a combination of several depth-wise separable convolutions. ShiftNets~\cite{shiftnet} proposed to use shift operation as an alternative to 3-by-3 convolution. AddressNets~\cite{addressnet} proposed three shift-based primitives for further improving performance on GPUs (Graphics Processing Units).

\subsection{Sparse Connection}
Shown in Figure~\ref{fig:related}~(b), \sparseconnect\ eliminates connections between neurons~\cite{han2015learning,liu2015sparse,lebedev2015fast,han2016eie,guo2016dynamic}.
XNOR-Net~\cite{Rastegari2016} binarized the connections. \cite{yang2016designing} prunes connections based on weights magnitude.   Deep compression~\cite{han2015deep} could accelerate fully connected layers up to \x{50}.
However, in practice, the actual speed-up may be strongly related to implementation. The standard library like CUDNN~\cite{cudnn} does not support sparse convolution very well. On the contrary, large matrices multiplication is highly optimized.

\subsection{Channel Pruning}
Shown in Figure~\ref{fig:related}~(c), \channelpruning\ aims at removing inter-channel redundancies of feature maps.
There were several training-based approaches:
\cite{Alvarez2016,wen2016learning,zhou2016less} regularize networks to improve accuracy.
Channel-wise SSL~\cite{wen2016learning} reaches high compression ratio for the first few convolutional layers of LeNet~\cite{lecun1998gradient} and AlexNet~\cite{krizhevsky2012imagenet}.
\cite{zhou2016less} could work well for fully connected layers.
However, \textit{training-based} approaches are more costly, and their effectiveness on very deep networks on large datasets is rarely exploited. 

Inference-time \channelpruning\ is challenging, as reported by previous works~\cite{anwar2015structured,polyak2015channel}. Channel pruning~\cite{cp} proposed to prune neural networks layer-by-layer using LASSO regression and linear least square reconstruction. Some works~\cite{srinivas2015data,mariet2015diversity,hu2016network} focus on model size compression, which mainly operate the fully connected layers. Data-free approaches~\cite{Li2016,anwar2016compact} results for \ratio\ (\eg, \x{5}) have not been reported, and requires long retraining procedure. \cite{anwar2016compact} select channels via over 100 random trials. However, it needs a long time to evaluate each trial on a deep network, which makes it infeasible to work on very deep models and large datasets. Recently, AMC~\cite{adc,amc} improves general channel pruning approach by learning the speed-up ratio with reinforcement learning. 

\subsection{Tensor Decomposition}
\Tensordecom\ methods~\cite{jaderberg2014speeding,lebedev2014speeding,gong2014compressing,kim2015compression} aim to approximate the original convolutional layer weights with several pieces of decomposed weights. Shown in Figure~\ref{fig:related}~(d), spatial decomposition~\cite{jaderberg2014speeding} factorized a layer into $3\times1$ and $1\times3$ combination, driven by spatial feature map redundancy. Shown in Figure~\ref{fig:related}~(e), channel decomposition~\cite{cd} factorized a layer into $3\times3$ and $1\times1$ combination, driven by channel-wise feature map redundancy.  \cite{xue2013restructuring,denton2014exploiting,fast} accelerate fully connected layers with truncated SVD. 

A simple hypothesis is that there is no need to decompose neural networks, simply just train them from scratch. In \cite{cd}, it has been empirically verified that decomposing neural networks can achieve better performance than naively training them from scratch. This motivates us to decompose regular convolution into \sepconv, which could potentially improve the performance of current compact neural networks which favor \sepconv\ a lot.  

\subsection{Implementation based acceleration}
Though convolution is a well defined operation, the run-time can largely depend on the implementations. \Implementation\ based methods~\cite{mathieu2013fast,vasilache2014fast,lavin2015fast,bagherinezhad2016lcnn} accelerate convolution, with special convolution algorithms like FFT~\cite{vasilache2014fast}.
Quantization~\cite{courbariaux2016binarynet,Rastegari2016} reduces floating point computational complexity, which is usually followed by fine-tuning and Huffman coding~\cite{han2015deep}. BinaryNet~\cite{courbariaux2016binarynet,mobinet} proposed to binarize both connections and weights. Recently, HAQ~\cite{haq} automated this process, which further compressed deep neural networks. These methods also depend on the hardware and library implementation.

\begin{figure*}
    \begin{center}
      \includegraphics[width=\linewidth]{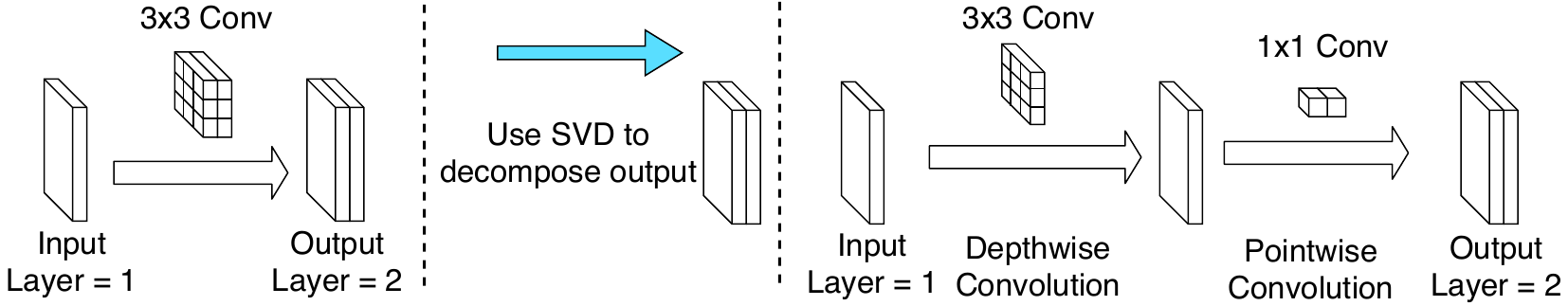}
    \end{center}
    \caption{\ourapproach\ for expanding regular convolution into \sepconv 
}
    \label{fig:approach}
\end{figure*}

\section{Approach}

In this section, we first introduce our approach for a special case: single channel convolutional layer. Then we generalize our approach to multi-channel convolutional layer case. Finally, we further generalize our approach for the multi-channel convolutional layer and multi-layers neural networks case.

Let $n$ and $c$ be the number of output and input channels. Let $N$ be the number of samples. Let $k_h$ and $k_w$ be the kernel size. Let $H$ and $W$ be the spatial size of the feature maps. Formally, we consider a regular convolutional layer shown in the left part of Figure~\ref{fig:approach}, where the weights tensor $W$ ($n\times c\times k_h\times k_w$) is applied to the input feature maps ($N\times c\times H\times W$) which results in the output feature maps ($N\times n\times H\times W$). The weights tensor can be decomposed into depth-wise convolutional weights $D$ and point-wise convolutional weights $P$ with shape $c\times c\times k_h\times k_w$ and $n\times c\times 1\times 1$ respectively,  shown in the right part of Figure~\ref{fig:approach}. In the following formulations, we do not consider the bias term for simplicity.

\subsection{Data Processing}
As we can observe, the spatial size $H$ and $W$ are usually large. Following channel decomposition~\cite{cd}, in this paper, we only consider sample patches $Y$ of size ($N\times n$) from the output feature maps and the corresponding $X$ of size ($N\times ck_hk_w$) from the input feature maps. The spatial size of the $Y$ patches is $1\times1$. We observe that it is enough to randomly sample 10 patches per image and 300 images in total from the training data to perform our decomposition (In this case, $N=300\times 10$). Convolution can be represented as matrix multiplication in our context:

\aeq{Y = XW}

W is the weights matrix ($ck_h k_w\times n$) reshaped from the original weights tensor.

\subsection{A Brief Introduction of Channel Decomposition}
Driven by channel-wise feature map redundancy, channel decomposition~\cite{cd} factorized a convolutional layer ($n\times c\times k_h\times  k_w$) into two convolutional layers: $W_1$ ($c'\times c\times k_h\times  k_w$) and $W_2$ ($n\times c'\times 1\times 1$) combination. This is achieved by finding a projection vector $P$ ($n\times c'$) such that:
\aeq{\argmin_{V} \lVert Y - YPP^T\rVert}
This problem can be solved by SVD:
\aeq{U,S,V&=SVD(Y)}
$P$ is the first $c'$ columns of matrix $V$. Then we can obtain $W_1, W_2$ as follow:
\aeq{W_1 &= PW \\
W_2 &= P \\}
\subsection{Single Channel Case}
In this case, we only have a single channel therefore the shape of the layer is $n\times 1\times H\times W$. We aim to decompose the weights tensor ($n\times 1\times k_h\times k_w$) into a depth-wise tensor $D$ and a point-wise tensors $P$ with shape $1\times 1\times k_h\times k_w$ and $n\times 1\times 1\times 1$ respectively. Let the output tensor $Y$ be the randomly sampled patches ($N\times n$). Let $V_0$ be the projection vector (size $n$). Formally, we try to find $V_0$ that minimize:
\aeq{\argmin_{V_0} \lVert Y - YV_0V_0^T\rVert}
SVD is employed to decompose the output tensor $Y$:
\aeq{U,S,V&=SVD(Y)}
To construct the two depth-wise and point-wise tensors  $D$ and $P$, we extract the projection vector $V_0$ (size $n$), namely first column of $V$:
\aeq{
D &= V_0W \\
P &= V_0 \\
}\label{eq:v0}
Then $D$ and $P$ are reshaped to the tensor format.

\subsection{Multi-Channel Case}
In single channel case, each channel is approximated into one depth-wise convolutional layer and one point-wise convolutional layer. We can simply apply the single channel algorithm multiple time to each channel of a multi-channel convolutional layer, as is also illustrated in Figure~\ref{fig:approach}:
\begin{algorithm}[H]
\caption {Depth-wise Decomposition}{$X_i$ is the $i$th channel of feature maps (shape: $N\times k_h k_w$). $W_i$ is the weights connected to $i$th channel (shape: $k_h k_w\times n$).}\label{alg:dd}
\begin{algorithmic}
\State {initialize D to a zero tensor}
\State {initialize P to a zero tensor}
\For{$i \gets 1$ to $c$}    
\State {$Y_i = X_iW_i$}
\State {$U,S,V=SVD(Y_i)$}
\State {$D_i = V_0W_i$}
\State {$P_i = V_0$}
\EndFor
    \State \Return {$D$, $P$}
\end{algorithmic}
\end{algorithm}

\subsection{Multi-Channel Case with Inter-channel Error Compensation}
\label{sec:mcc}
Approximation in multi-channel, however, can lead to large approximation error. Thus we propose a novel way to decompose multi-channel sequentially taking into account the approximation errors introduced by decomposition of previous channels (Experiments in Section~\ref{sec:single}). Specifically, instead of purely approximating the output tensor $Y_i$, we approximate both the output tensor $Y_i$ and approximation error introduced by other channels $E$:
\aeq{\argmin_{V_0} & \lVert Y_i + E - Y_iU_0V_0^T\rVert \\
E &= \sum_{k=0}^{i-1} E_k\\
E_i &= \lvert Y_i - Y_iU_0V_0^T \rvert\\
}
where $\lvert\cdot\rvert$ is the absolute value function. 

This can be solve by  Generalized Singular
Value Decomposition (GSVD)~\cite{gsvd}, without the need of stochastic gradient descent (SGD):
\aeq{U,S,V&=GSVD(Y_i + E, Y_i)\\
D_i &= W_iU_0\\
P_i &= S_0V_0\\}
Similar with Equation~\ref{eq:v0}, $U_0$ and $V_0$ are the first columns of matrices $U$ and $V$ respectively. 

The \ourapproach\  with inter-channel error compensation  algorithm is as follow:
\begin{algorithm}[H]
\caption {Depth-wise Decomposition with compensation}{$X_i$ is the $i$th channel of feature maps (shape: $N\times k_h k_w$). $W_i$ is the weights connected to $i$th channel (shape: $k_h k_w\times n$). $E$ is the tensor of accumulated error.}\label{alg:dd}
\begin{algorithmic}
\State {initialize D to a zero tensor}
\State {initialize P to a zero tensor}
\State {initialize E to a zero tensor}
\For{$i \gets 1$ to $c$}    
\State {$Y_i = X_iW_i$}
\State {$U,S,V=GSVD(Y_i + E, Y_i)$}
\State {$D_i = W_iU_0$}
\State {$P_i = S_0V_0$}
\State {$E_i = Y_i - Y_iU_0V_0^T$}
\State {$E = E + E_i$}
\EndFor
    \State \Return {$D$, $P$}
\end{algorithmic}
\end{algorithm}

\subsection{Multi-layer Case}\label{sec:multlayercase}
Finally, our approach can be applied to deep convolutional neural networks layer-by-layer (Section~\ref{sec:multi}). Similar with Section~\ref{sec:mcc}, we can still account for the accumulated error introduced by multi-layer approximation. Here, the error $E^l$ in layer $l$ includes both approximation error introduced by decomposing other channels and the previous layer $l-1$. We simply need to extract all feature maps patches $Y_i'$ before doing decomposition, since the feature map responses will change during decomposition. Similar to the previous sub-section, this problem can be solved by GSVD:
\aeq{\argmin_{V_0} &\lVert Y_i' + E^l - Y_iU_0V_0^T\rVert\\
E^l &= E^{l-1} + \sum_{k=0}^{i-1} E_k\\
E_i &= \lvert Y_i' - Y_iU_0V_0^T \rvert\\
U,S,V&=GSVD(Y_i' + E^l, Y_i)}

The following algorithm explains the detailed procedure:
\begin{algorithm}[H]
\caption {Multi-layer Decomposition with compensation}{$X_i$ is the $i$th channel of feature maps (shape: $N\times k_h k_w$). $X_i'$ is the ground truth $i$th channel of feature maps before decomposing the network. $W_i$ is the weights connected to $i$th channel (shape: $k_h k_w\times n$). $E$ is the tensor of accumulated error.}\label{alg:dd}
\begin{algorithmic}
\State {initialize D to a zero tensor}
\State {initialize P to a zero tensor}
\If{$l==0$}
\State {initialize $E^l$ to a zero tensor}
\Else
\State {initialize $E^l$ to the error of previous layer $E^{l-1}$}
\EndIf
\For{$i \gets 1$ to $c$}    
\State {$Y_i' = X_i'W_i$}
\State {$Y_i = X_iW_i$}
\State {$U,S,V=GSVD(Y_i' + E, Y_i)$}
\State {$D_i = W_iU_0$}
\State {$P_i = S_0V_0$}
\State {$E_i = Y_i' - Y_iU_0V_0^T$}
\State {$E^l = E^l + E_i$}
\EndFor
    \State \Return {$D$, $P$}
\end{algorithmic}
\end{algorithm}

\subsection{Fine-tuning}\label{sec:finetune}
Following channel decomposition~\cite{cd}, after a deep convolutional neural network is fully decomposed, we can fine-tune the model for ten epochs with a small learning rate $1e^{-4}$ to obtain better accuracy (Section~\ref{sec:whole}). 

Unless specified, we do not fine-tune the decomposed model in the following experiments.

\section{Experiments}
We conduct rich experiments on ImageNet~\cite{imagenet} 2012 classification dataset. ImageNet is a very large scale image classification dataset which consists of 1000 classes. Our ShuffleNet V2 model~\cite{shufflenetv2}
is trained on the 1.3 million training images with 8 GeForce GTX TITAN X GPUs and CUDA 10~\cite{cuda} CUDNN 7.4~\cite{cudnn}. We also trained a folded ShuffleNet V2 model~\cite{shufflenetv2}. In the folded ShuffleNet V2 model~\cite{shufflenetv2} we replace every pair of a depthwise convolutional layer and pointwise convolutional layer with a regular convolutional layer.  Our model is evaluated on the 50,000 validation images. We evaluate the performance of our approach with top-1 error rate and relative error on ImageNet dataset.

Our neural networks implementation is based on TensorFlow~\cite{tensorflow}. Our implementation of the previous approach: channel decomposition~\cite{cd} is based on pure Python\footnote{github.com/yihui-he/channel-pruning}.

\begin{table}[]
\begin{center}
\begin{tabular}{cc}
\hline
\multicolumn{2}{c}{single layer $9\times$ acceleration with random data} \\ \hline\hline
\multicolumn{1}{c|}{} & relative error \\ \hline
\multicolumn{1}{c|}{Channel Decomposition~\cite{cd}} & $0.887\pm 1.34e^{-6}$ \\ \hline
\multicolumn{1}{c|}{Depth-wise Decomposition (ours)} & $0.914\pm3.21e^{-7}$ \\ \hline
\multicolumn{1}{c|}{Depth-wise Decomposition (ours)} & $0.906\pm2.78e^{-7}$ \\ 
 \multicolumn{1}{c|}{with compensation} &  \\ \hline
\end{tabular}
\end{center}
\caption{Sanity check. Using random data, our approach performs as good as channel decomposition~\cite{cd}. Standard deviations and relative errors are obtained after 10 runs}
\label{table:sanity}
\end{table}

\subsection{Sanity Check}
To check the correctness of our implementation, we created a random weights matrix of size $64\times128$ and the corresponding random input feature of size $3000\times64$. The resulting output response is, therefore, a $3000\times128$ matrix. Our baseline is channel decomposition~\cite{cd}.
Our \ourapproach\ decomposes a convolutional layer into a depth-wise convolution followed by a point-wise convolution, which accelerates the convolutional layer by around \x{9}. For a fair comparison, we use channel decomposition under \x{9} acceleration.

To measure the correctness of our approach, we measure the relative error of the resulting output responses between those of decomposing algorithms(including baseline~\cite{cd}, \ourapproach\, and \ourapproach\  with inter-channel compensation) and the ground-truth output response. 

As is shown in Table~\ref{table:sanity}, our \ourapproach\ is comparable to channel decomposition~\cite{cd}. 

We further tested the reconstruction error of \ourapproach\ with inter-channel error compensation. As expected, the relative error of this method is smaller than that of our basic approach.
\begin{figure}[h]
    \includegraphics[width=0.5\textwidth]{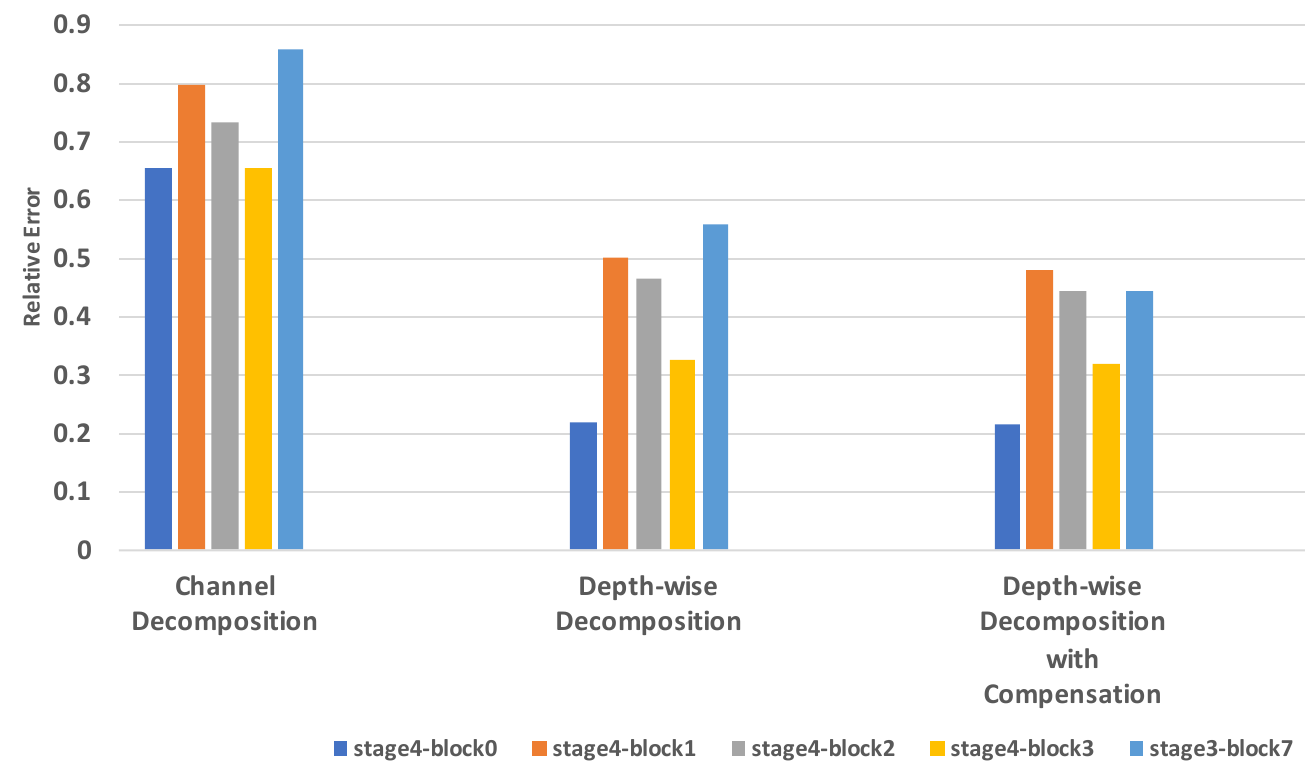}
    \caption{Relative error for decomposing a single conv layer in ShuffleNet V2 
}
    \label{fig:relative_error}
\end{figure}

\begin{figure*}
\begin{center}
\includegraphics[width=.49\textwidth]{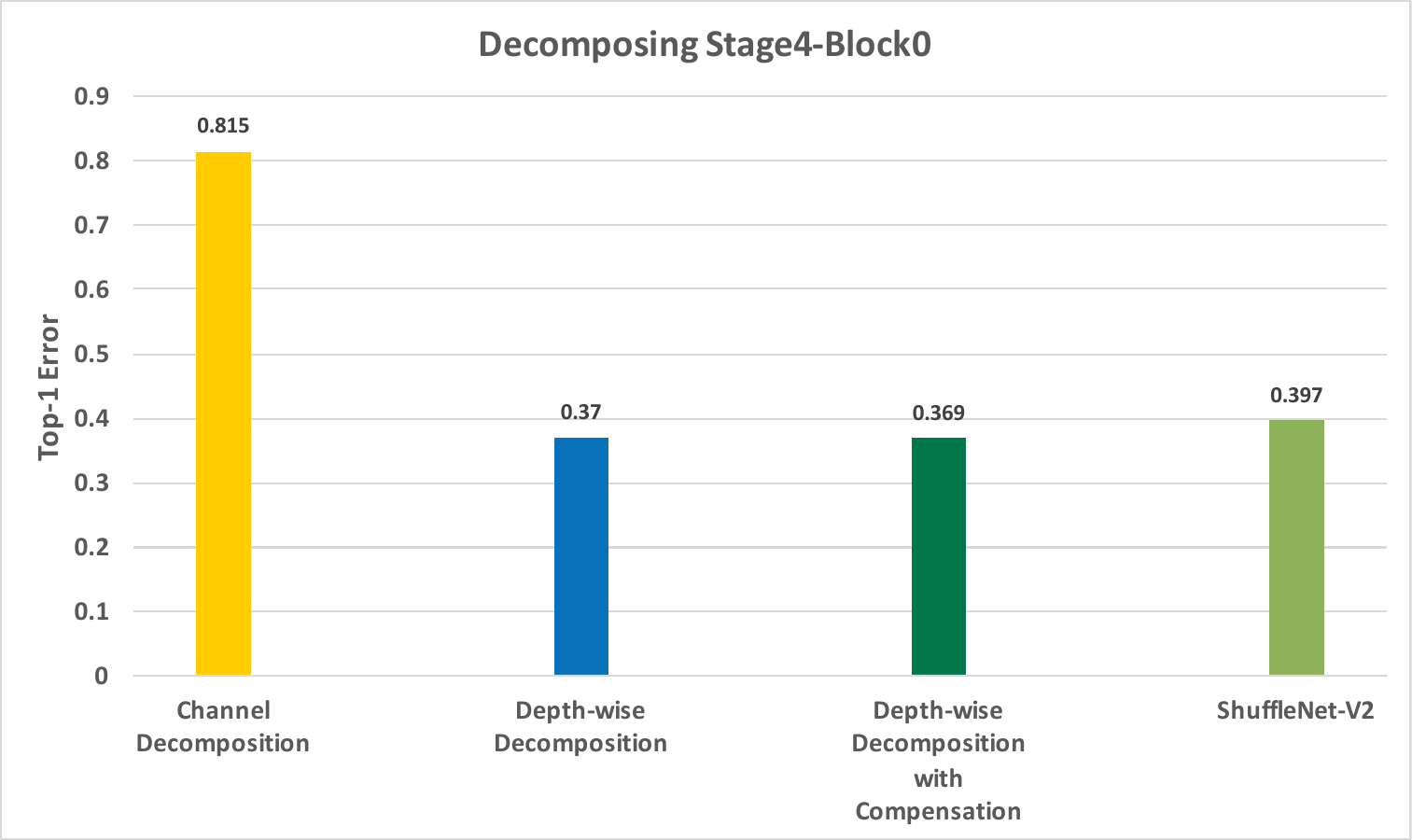}
\includegraphics[width=.49\textwidth]{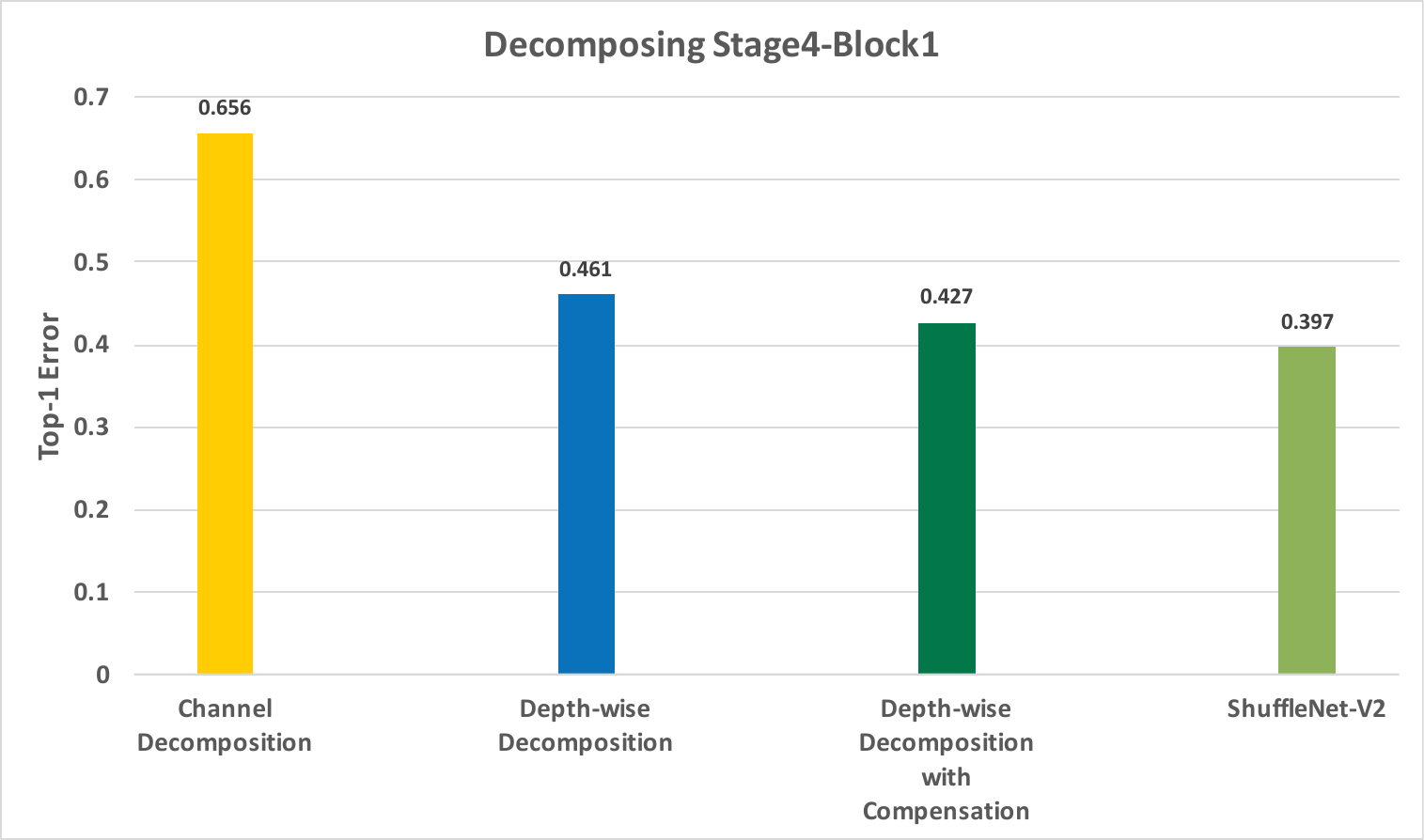}
\end{center}

\begin{center}
\includegraphics[width=.49\textwidth]{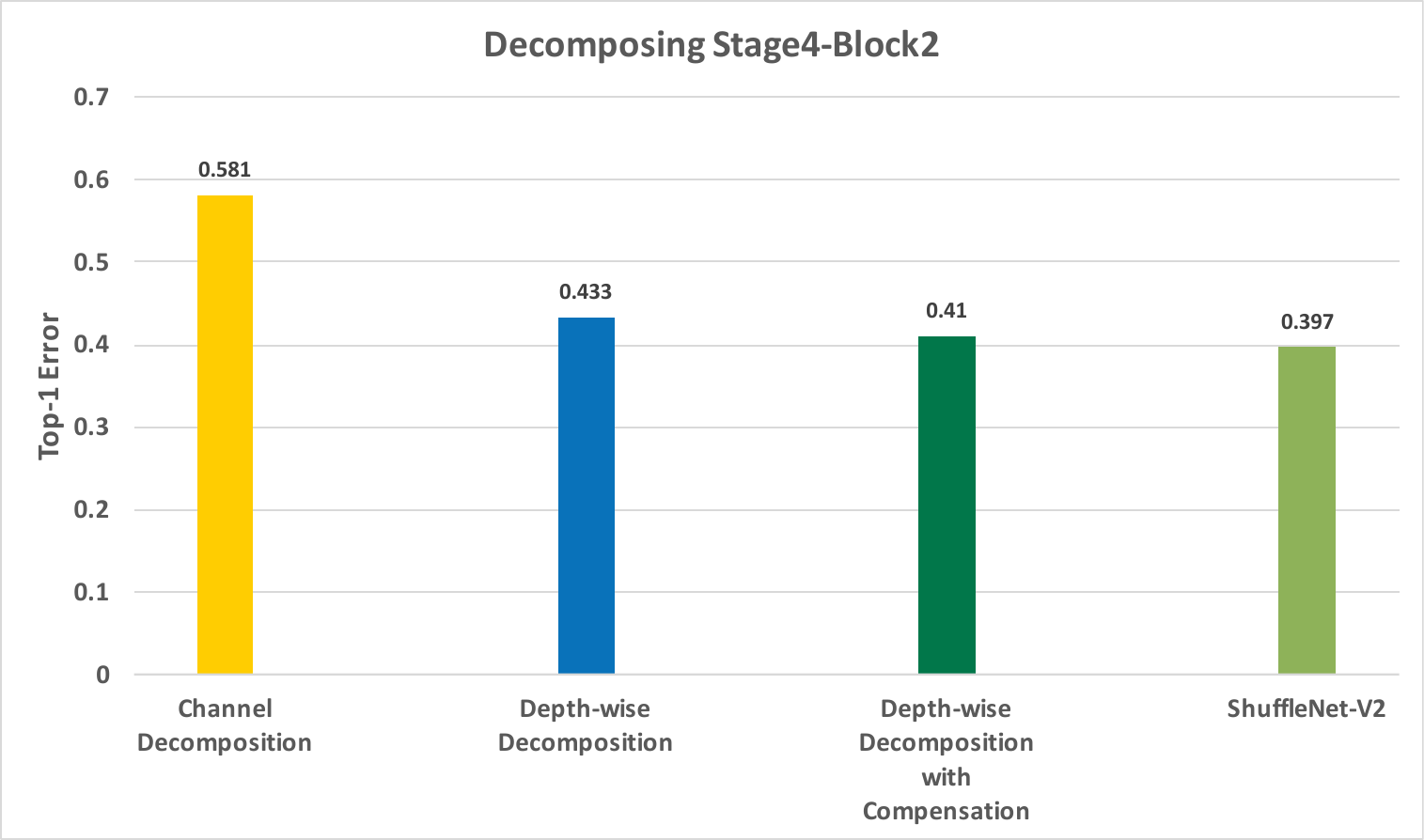}
\includegraphics[width=.49\textwidth]{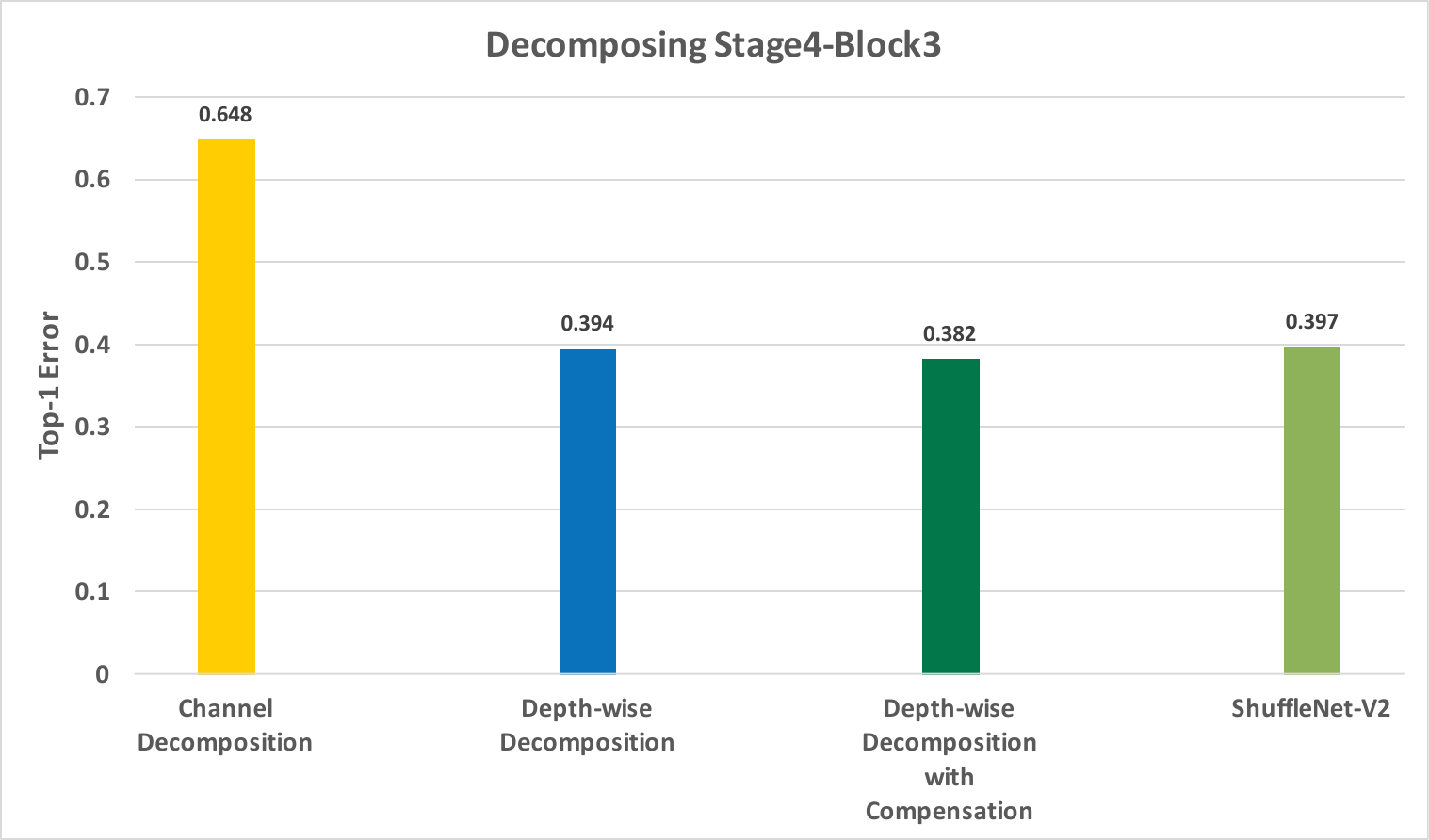}
\end{center}

    \caption{Top-1 testing error for decomposing a single convolutional layer in ShuffleNet V2~\cite{shufflenetv2}}
\label{fig:top1}
\end{figure*}

\begin{table}[]
\begin{center}
\begin{tabular}{cc}
\hline
\multicolumn{2}{c}{Whole Model Decomposition with ImageNet} \\ \hline\hline
\multicolumn{1}{c|}{} & top-$1$ error \\ \hline
\multicolumn{1}{c|}{ShuffleNet V2~\cite{shufflenetv2}} & 39.7\% \\ \hline
\multicolumn{1}{c|}{Channel Decomposition} & 40.0\% \\ 
 \multicolumn{1}{c|}{ with fine-tuning~\cite{cd} (our impl.)} &  \\ \hline
\multicolumn{1}{c|}{Folded ShuffleNet V2~\cite{shufflenetv2}} & 36.6\% \\ \hline
\multicolumn{1}{c|}{Depth-wise Decomposition} & 43.9\% \\ 
 \multicolumn{1}{c|}{with compensation (ours)} &  \\ \hline
\multicolumn{1}{c|}{Fine Tuned Depth-wise Decomposition} & \textbf{37.9\%} \\ 
 \multicolumn{1}{c|}{with compensation (ours)} &  \\ \hline
\end{tabular}
\end{center}
\caption{Comparisons on compressing ShuffleNet V2 0.5x}
\label{table:result}
\end{table}

\subsection{Single Layer Decomposition}\label{sec:single}
Firstly, we want to evaluate the reconstruction error of \ourapproach\ for a single layer. We conduct experiments on decomposing five different convolutional layers of ShuffleNet V2~\cite{shufflenetv2}. We decompose each 3-by-3 convolutional layer with the baseline method~\cite{cd}, \ourapproach\ and \ourapproach\ with error compensation.  Figure~\ref{fig:relative_error} shows the relative error of reconstructing each of the five convolutional layers. As shown in the figure, \ourapproach\ results in lower reconstruction error for all five layers. Furthermore, the \ourapproach\ with inter-channel compensation results in even smaller reconstruction error. This proves that both \ourapproach\ and \ourapproach\ with inter-channel error compensation are better at preserving the accuracy of the network while achieving the same level of effectiveness in terms of accelerating the network.

Secondly, we want to measure the effectiveness of \ourapproach. Thus we measure how much the top-1 error of the resulting network has increased when decomposing a single convolutional layer in ShuffleNet V2~\cite{shufflenetv2}. We demonstrate the effectiveness of \ourapproach\ by showing the top-1 error of the networks resulting from decomposing each of the four convolutional layers in stage 4 of ShuffleNet V2~\cite{shufflenetv2} along with the top-1 error achieved by the original ShuffleNet V2~\cite{shufflenetv2} model. As shown in Figure ~\ref{fig:top1}, the baseline method~\cite{cd}, when accelerating one convolutional layer by around \x{9}, largely increases the top-1 error of the resulting network. In contrast, \ourapproach, along with \ourapproach\ with error compensation do not have such a significant impact on the top-1 error of the resulting network.

\subsection{Whole Model Decomposition}\label{sec:whole}
Lastly we want to measure the effect of decomposing the whole model. First we train a "folded" ShuffleNet V2~\cite{shufflenetv2}. In this model we replace depth-wise separable convolutional layers with regular convolutional layers. For example, a depth-wise convolution layer with shape $C\times C\times k_h \times k_w$ and a point-wise convolution with shape $N\times C\times 1 \times 1$ will be replaced by a regular convolutional layer with shape $N\times C\times k_h \times k_w$. We train this model with the exact same settings as the original ShuffleNets V2 paper. Our implementation is based on TensorPack~\cite{tensorpack}\footnote{github.com/tensorpack/tensorpack}. Shown in Table~\ref{table:result}, the Top-1 accuracy of "folded" ShuffleNet V2 is \textbf{3.1\%} higher than the original ShuffleNet V2, which serves as the upper bound of our proposed algorithm. 

Then we decompose each convolutional layers in the folded ShuffleNet V2~\cite{shufflenetv2}  using our multi-layer channel decomposition method with inter-channel error compensation (Section~\ref{sec:multlayercase}). This decomposed model has the \textbf{exact same} architecture as the original ShuffleNet V2. Then the decomposed model is further fine-tuned on ImageNet training dataset for 10 epochs (Section~\ref{sec:finetune}). 

As a baseline, we perform channel decomposition~\cite{cd} under $9\times$ acceleration ratio on ShuffleNet V2 and fine-tune for the same number of epochs as our model. However, channel decomposition does not work well for high acceleration ratio in our case. Shown in Table~\ref{table:result}, it even performs worse than the original ShuffleNet V2.

As is shown in Table~\ref{table:result}, our \ourapproach\ with inter-channel error compensation has \textbf{$\sim$ 2\%} lower top-1 error than the original ShuffleNet V2~\cite{shufflenetv2}. It proves that our method is able to achieve a better level of accuracy under the same computational complexity compare to  ShuffleNet V2~\cite{shufflenetv2}.

\begin{table}[]
\begin{center}
\begin{tabular}{c|c|c}
\hline
top-$1$ error & baseline & ours \\ \hline
ShuffleNet V2 0.5$\times$~\cite{shufflenetv2} & 39.7\% & \textbf{37.9\%} \\ \hline
ShuffleNet V2 1.0$\times$~\cite{shufflenetv2} & 30.6\% & \textbf{28.0\%} \\ \hline
ShuffleNet V2 2.0$\times$~\cite{shufflenetv2} & 25.1\% & \textbf{23.4\%} \\ \hline
Xception~\cite{xception} & 21.0\% & \textbf{20.1\%} \\ \hline
\end{tabular}
\end{center}
\caption{Top-1 error for compressing ShuffleNet V2 family and Xception}
\label{table:all}
\end{table}

To test the generalizability of our approach, we further test on other ShuffleNet V2 architectures and Xception. Shown in Table~\ref{table:all}, The results are consistent with ShuffleNet v2 $0.5\times$ in Table~\ref{table:result}. For ShuffleNet v2 $1\times$, our approach improves the Top-1 accuracy by \textbf{1.6\%}. For ShuffleNet v2 $2\times$, the Top-1 accuracy improvement is \textbf{1.7\%}. For Xception, the model performs \textbf{1.6\%} better.

\section{Conclusion}
In conclusion, very deep convolutional neural networks (CNNs) are widely used among many computer vision tasks. However, most state-of-the-art CNNs are large, which results in high inference latency. Recent depth-wise separable convolution has been proposed for image recognition tasks on the computationally limited platforms. Though it is much faster than regular convolution, accuracy is sacrificed. In this paper, we propose a novel decomposition approach based on SVD, namely \approach, for expanding regular convolution into depth-wise separable convolution while maintaining high accuracy. Thorough experiments with the latest ShuffleNet V2 model~\cite{shufflenetv2} on ImageNet~\cite{imagenet} have verified the effectiveness of our approach.

\nocite{kumar2019pack}
{\small
\bibliographystyle{ieee}
\bibliography{egbib}
}

\end{document}